\documentclass[11pt,a4paper]{article}
\usepackage[hyperref]{acl2021}
\usepackage{times}
\usepackage{latexsym}

\usepackage{microtype}

\usepackage{graphicx}
\usepackage{subcaption}

 \aclfinalcopy 
\def\aclpaperid{1372} 

\title{Including Signed Languages in Natural Language Processing}

\author{Kayo Yin$^1$, Amit Moryossef$^2$, Julie Hochgesang$^3$, Yoav Goldberg$^{2,4}$, Malihe Alikhani$^5$ \\
$^1$Language Technologies Institute, Carnegie Mellon University\\
$^2$Bar-Ilan University\\
$^3$Department of Linguistics, Gallaudet University\\
$^4$Allen Institute for AI\\
$^5$School of Computing and Information, University of Pittsburgh\\
\texttt{kayoy@cs.cmu.edu, amitmoryossef@gmail.com}\\ 
\texttt{julie.hochgesang@gallaudet.edu, yogo@cs.biu.ac.il, malihe@pitt.edu}}

\date{}

\begin{document}
\maketitle
\begin{abstract}
Signed languages are the primary means of communication for many deaf and hard of hearing individuals. Since signed languages exhibit all the fundamental linguistic properties of natural language, we believe that tools and theories of Natural Language Processing (NLP) are crucial towards its modeling. However, existing research in Sign Language Processing (SLP) seldom attempt to explore and leverage the linguistic organization of signed languages. This position paper calls on the NLP community to include signed languages as a research area with high social and scientific impact. We first discuss the linguistic properties of signed languages to consider during their modeling. Then, we review the limitations of current SLP models and identify the open challenges to extend NLP to signed languages. Finally, we urge (1) the adoption of an efficient tokenization method; (2) the development of linguistically-informed models; (3) the collection of real-world signed language data;  (4) the inclusion of local signed language communities as an active and leading voice in the direction of research. 
\end{abstract}

\section{Introduction}

Natural Language Processing (NLP) has revolutionized the way people interact with technology through the rise of personal assistants and machine translation systems, to name a few. 
However, the vast majority of NLP models require a spoken language input (speech or text), thereby excluding around 200 different signed languages and up to 70 million deaf people\footnote{According to World Federation of the Deaf\\ \url{https://wfdeaf.org/our-work/}} from modern language technologies.

\begin{figure}[t]
    \centering
    \includegraphics[width=\linewidth]{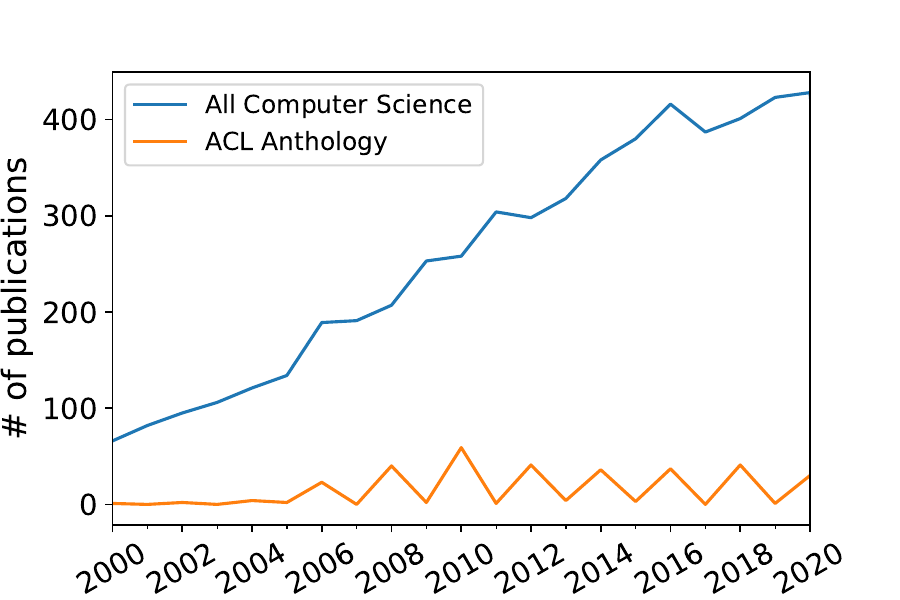}
    \caption{Evolution of the number of publications referring to sign language in its title from computer science venues and in the ACL anthology. Publications in computer science are extracted from the Semantic Scholar archive \cite{ammar-etal-2018-construction}.}
    \label{fig:ss-years}
\end{figure}

Throughout history, Deaf communities fought for the right to learn and use signed languages, as well as for the recognition of signed languages as legitimate languages (\S\ref{sec:bg}). Indeed, signed languages are sophisticated communication modalities that are at least as capable as spoken languages in all manners, linguistic and social. 
However, in a predominantly oral society, deaf people are constantly encouraged to use spoken languages through lip-reading or text-based communication.
The exclusion of signed languages from modern language technologies further suppresses signing in favor of spoken languages.  This disregards the preferences of the Deaf communities who strongly prefer to communicate in signed languages both online and for in-person day-to-day interactions, among themselves and when interacting with spoken language communities \cite{padden1988deaf, glickman2018language}. Thus, it is essential to make signed languages accessible.

To date, a large amount of research on Sign Language Processing (SLP) has been focused on the visual aspect of signed languages, led by the Computer Vision (CV) community, with little NLP involvement (Figure \ref{fig:ss-years}). This is not unreasonable, given that a decade ago, we lacked the adequate CV tools to process videos for further linguistic analyses. However, like spoken languages, signed languages are fully-fledged systems that exhibit all the fundamental characteristics of natural languages (\S\ref{sec:linguistics}), and current SLP techniques fail to address or leverage the linguistic structure of signed languages (\S\ref{sec:modeling}). This leads us to believe that NLP tools and theories are crucial to process signed languages.
Given the recent advances in CV, this position paper argues that now is the time to incorporate linguistic insight into signed language modeling.

Signed languages introduce novel challenges for NLP due to their visual-gestural modality, simultaneity, spatial coherence, and lack of written form. By working on signed languages, the community will gain a more holistic perspective on natural languages through a better understanding of how meaning is conveyed by the visual modality and how language is grounded in visuospatial concepts. 

Moreover, SLP is not only an intellectually appealing area but also an important research area with a strong potential to benefit signing communities.  
Examples of beneficial applications enabled by signed language technologies include better documentation of endangered sign languages; educational tools for sign language learners; tools for query and retrieval of information from signed language videos; personal assistants that react to signed languages; real-time automatic sign language interpretations, and more.
Needless to say, in addressing this research area, researchers should work \emph{alongside} and \emph{under the direction of} deaf communities, and to the benefit of the signing communities' interest above all \cite{harris2009research}.

After identifying the challenges and open problems to successfully include signed languages in NLP (\S\ref{sec:applications}), we emphasize the need to: (1) develop a standardized tokenization method of signed languages with minimal information loss for its modeling; (2) extend core NLP technologies to signed languages to create linguistically-informed models; (3) collect signed language data of sufficient size that accurately represents the real world; (4) involve and collaborate with the Deaf communities at every step of research.

\section{Background and Related Work}
\label{sec:bg}

\subsection{History of Signed Languages and Deaf Culture}

Over the course of modern history, spoken languages were dominant so much so that signed languages struggled to be recognized as languages in their own right and educators developed misconceptions that signed language acquisition may hinder the development of speech skills. For example, in 1880, a large international conference of deaf educators called the \textit{Second International Congress on Education of the Deaf} banned teaching signed languages, favoring speech therapy instead. It was not until the seminal work on American Sign Language (ASL) by \citet{stokoe} that signed languages started gaining recognition as natural, independent, and well-defined languages, which then inspired other researchers to further explore signed languages as a research area.
Nevertheless, antiquated notions that deprioritized signed languages continue to do harm and subjects many to linguistic neglect \cite{humphries2016avoiding}. Several studies have shown that deaf children raised solely with spoken languages do not gain enough access to a first language during their critical period of language acquisition \cite{murray2020importance}. This language deprivation can lead to life-long consequences on the cognitive, linguistic, socioemotional, and academic development of the deaf \cite{hall2017language}. 

Signed languages are the primary languages of communication for the Deaf\footnote{When capitalized, ``Deaf'' refers to a community of deaf people who share a language and a culture, whereas the lowercase ``deaf'' refers to the audiological condition of not hearing.} and are at the heart of Deaf communities. Failing to recognize signed languages as fully-fledged natural language systems in their own right has had harmful effects in the past, and in an increasingly digitized world, the NLP community has an important responsibility to include signed languages in its research. NLP research should strive to enable a world in which all people, including the Deaf, have access to languages that fit their lived experience.

\subsection{Sign Language Processing in the Literature}

\citet{jaffe1994evolution, ong2005automatic, parton2006sign} survey early works in SLP that were mostly limited to using sensors to capture fingerspelling and isolated signs, or use rules to synthesize signs from spoken language text, due to the lack of adequate CV technology at the time to process videos. This paper will instead focus on more recent vision-based and data-driven approaches that are non-intrusive and more powerful. 
The introduction of a continuous signed language benchmark dataset \cite{dataset:forster2014extensions,cihan2018neural}, coupled with the advent of deep learning for visual processing, lead to increased efforts to recognize signed expressions from videos. Recent surveys on SLP mostly review these different approaches for sign language recognition developed by the CV community \cite{koller2020quantitative,rastgoo2020sign,adaloglou2020comprehensive}. 

Meanwhile, signed languages have remained relatively overlooked in NLP literature (Figure \ref{fig:ss-years}). \citet{bragg-interdisciplinary} argue the importance of an interdisciplinary approach to SLP, raising the importance of NLP involvement among other disciplines. We take this argument further by diving into the linguistic modeling challenges for signed languages and providing a roadmap of open questions to be addressed by the NLP community, in hopes of stimulating efforts from an NLP perspective towards research on signed languages. 
Conversely, \citet{moryossef2021slp} organize the sign language processing literature, datasets, and tasks; however, they do not suggest and actionable research directions.



\section{Sign Language Lingusitics}\label{sec:linguistics}

Signed languages consist of phonological, morphological, syntactic, and semantic levels of structure that fulfill the same social, cognitive, and communicative purposes as other natural languages. While spoken languages primarily channel the oral-auditory modality, signed languages use the visual-gestural modality, relying on the face, hands, body of the signer, and the space around them to create distinctions in meaning. We present the linguistic features of signed languages\footnote{We mainly refer to ASL, where most sign language research has been conducted, but not exclusively.} that must be taken into account during their modeling.
%

\paragraph{Phonology}
Signs are composed of minimal units that combine manual features such as hand configuration, palm orientation, placement, contact, path movement, local movement, as well as non-manual features including eye aperture, head movement, and torso positioning\footnote{In this work, we focus on visual signed languages rather than tactile systems such as Pro-Tactile ASL which DeafBlind Americans sometimes prefer.}  \cite{liddell1989american, johnson2011toward, brentari2011sign, sandler2012phonological}. In both signed and spoken languages, not all possible phonemes are realized, and inventories of two languages' phonemes/features may not overlap completely. Different languages are also subject to rules for the allowed combinations of features.

\paragraph{Simultaneity}
Though an ASL sign takes about twice as long to produce than an English word, the rates of transmission of information between the two languages are similar \cite{bellugi1972comparison}. One way signed languages compensate for the slower production rate of signs is through simultaneity: signed languages make use of multiple visual cues to convey different information simultaneously\cite{sandler2012phonological}. 
For example, the signer may produce the sign for 'cup' on one hand while simultaneously pointing to the actual cup with the other to express ``that cup''. 
Similarly to tone in spoken languages, the face and torso can convey additional affective information \cite{liddell2003grammar,johnston2007australian}. Facial expressions can modify adjectives, adverbs, and verbs; a head shake can negate a phrase or sentence; eye direction can help indicate referents.

\paragraph{Referencing}
The signer can introduce referents in discourse either by pointing to their actual locations in space, or by assigning a region in the signing space to a non-present referent and by pointing to this region to refer to it \cite{rathmann2011featural, schembri2018indicating}. Signers can also establish relations between referents grounded in signing space by using directional signs or embodying the referents using body shift or eye gaze \cite{dudis2004body, liddell1998gesture}. Spatial referencing also impacts morphology when the directionality of a verb depends on the location of the reference to its subject and/or object \cite{de2008pointing, fenlon2018modification}: for example, a directional verb can move from the location of its subject and ending at the location of its object. While the relation between referents and verbs in spoken language is more arbitrary, referent relations are usually grounded in signed languages. The visual space is heavily exploited to make referencing clear. 

Another way anaphoric entities are referenced in sign language is by using classifiers or depicting signs \cite{supalla1986classifier, wilcox2004rethinking, roy2011discourse} that help describe the characteristics of the referent. Classifiers are typically one-handed signs that do not have a particular location or movement assigned to them, or derive features from meaningful discourse \cite{liddell2003grammar}, so they can be used to convey how the referent relates to other entities, describe its movement, and give more details. For example, to tell about a car swerving and crashing, one might use the hand classifier for a vehicle, move it to indicate swerving, and crash it with another entity in space.

To quote someone other than oneself, signers perform \textit{role shift} \cite{cormier2015rethinking}, where they may physically shift in space to mark the distinction, and take on some characteristics of the people they are representing. For example, to recount a dialogue between a taller and a shorter person, the signer may shift to one side and look up when taking the shorter person's role, shift to the other side and look down when taking the taller person's role.

\begin{figure*}[ht!]
    \centering
    \includegraphics[width=\linewidth]{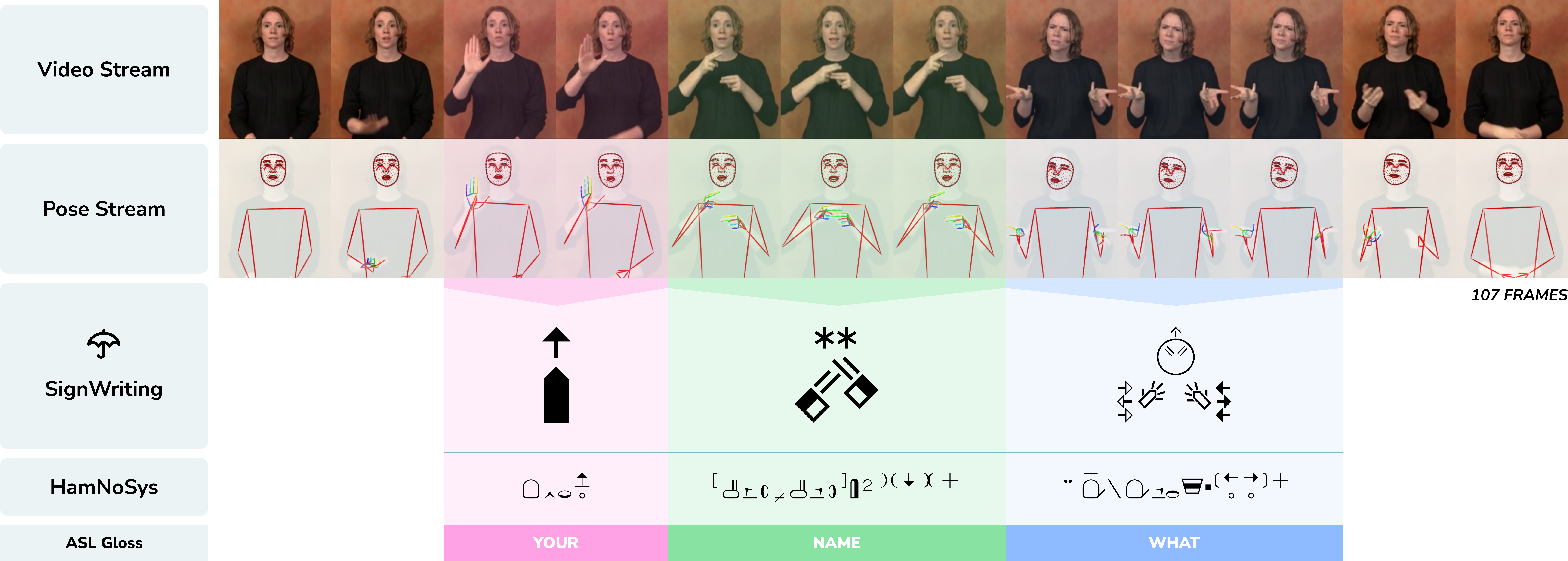}
    \caption{Representations of an American Sign Language phrase with video frames, pose estimations, SignWriting, HamNoSys and glosses. English translation: ``What is your name?''}
    \label{fig:modeling}
\end{figure*}

\paragraph{Fingerspelling}
Fingerspelling is a result of language contact between a signed language and a surrounding spoken language written form \cite{battison1978lexical,wilcox1992phonetics,brentari2001language, patrie2011fingerspelled}. A set of manual gestures correspond with a written orthography or phonetic system. Fingerspelling is often used to indicate names or places or new concepts from the spoken language but often have become integrated into the signed languages themselves as another linguistic strategy \cite{padden1998asl,montemurro2018emphatic}.

\section{Current State of SLP}\label{sec:modeling}


In this section, we present the existing methods, resources, and tasks in SLP, and discuss their limitations to lay the ground for future research. 

\subsection{Representations of Signed Languages}\label{sec:modeling-rep}
Representation is a significant challenge for SLP, as unlike spoken languages, signed languages have no widely adopted written form. Figure \ref{fig:modeling} illustrates each signed language representation we will describe below.

\paragraph{Videos} are the most straightforward representation of a signed language and can amply incorporate the information conveyed through sign. 
One major drawback of using videos is their high dimensionality: they usually include more information than needed for modeling, and are expensive to store, transmit, and encode. As facial features are essential in sign, anonymizing raw videos also remains an open problem, limiting the possibility of making these videos publicly available \cite{isard2020approaches}.

\paragraph{Poses} reduce the visual cues from videos to skeleton-like wireframe or mesh representing the location of joints. 
While motion capture equipment can often provide better quality pose estimation, it is expensive and intrusive, and estimating pose from videos is the preferred method currently \cite{pose:pishchulin2012articulated,pose:chen2017adversarial,pose:cao2018openpose,pose:alp2018densepose}.
Compared to video representations, \emph{accurate} poses are lower in complexity and anonymized, while observing relatively low information loss. However, they remain a continuous, multidimensional representation that is not adapted to most NLP models.

\paragraph{Written notation systems} represent signs as discrete visual features. Some systems are written linearly and others use graphemes in two dimensions.
While various universal \cite{writing:sutton1990lessons, writing:prillwitz1990hamburg} and language-specific notation systems \cite{writing:stokoe2005sign,writing:kakumasu1968urubu, writing:bergman1977tecknad} have been proposed, no writing system has been adopted widely by any sign language community, and the lack of standard hinders the exchange and unification of resources and applications between projects.
Figure \ref{fig:modeling} depicts two universal notation systems: SignWriting \cite{writing:sutton1990lessons}, a two-dimensional pictographic system, and HamNoSys \cite{writing:prillwitz1990hamburg}, a linear stream of graphemes that was designed to be readable by machines.

\paragraph{Glossing} is the transcription of signed languages sign-by-sign, where every sign has a unique identifier.
While various sign language corpus projects have provided gloss annotation guidelines \cite{mesch2015gloss,johnston2016auslan,konrad2018public}, again, there is no single agreed-upon standard. Linear gloss annotations are also an imprecise representation of signed language: they do not adequately capture all information expressed simultaneously through different cues (i.e. body posture, eye gaze) or spatial relations, which leads to an inevitable information loss up to a semantic level that affects downstream performance on SLP tasks \cite{yin-read-2020-better}.

\subsection{Existing Sign Language Resources}\label{sec:modeling-resources}
Now, we introduce the different formats of resources and discuss how they can be used for signed language modeling.

\paragraph{Bilingual dictionaries} for signed language \cite{dataset:mesch2012meaning,fenlon2015building,crasborn2016ngt,dataset:gutierrez2016lse} map a spoken language word or short phrase to a signed language video.
One notable dictionary is, SpreadTheSign\footnote{\url{https://www.spreadthesign.com/}} is a parallel dictionary containing around 23,000 words with up to 41 different spoken-signed language pairs and more than 500,000 videos in total. While dictionaries may help create lexical rules between languages, they do not demonstrate the grammar or the usage of signs in context.

\paragraph{Fingerspelling corpora} usually consist of videos of words borrowed from spoken languages that are signed letter-by-letter. They can be synthetically created \cite{dataset:dreuw2006modeling} or mined from online resources \cite{dataset:fs18slt,dataset:fs18iccv}. However, they only capture one aspect of signed languages.

\paragraph{Isolated sign corpora} are collections of annotated single signs. They are synthesized \cite{dataset:ebling2018smile,dataset:huang2018video,dataset:sincan2020autsl,dataset:hassan-etal-2020-isolated} or mined from online resources \cite{dataset:joze2018ms,dataset:li2020word}, and can be used for isolated sign language recognition or for contrastive analysis of minimal signing pairs \cite{dataset:imashev2020dataset}. However, like dictionaries, they do not describe relations between signs nor do they capture coarticulation during signing, and are often limited in vocabulary size (20-1000 signs)
 
\paragraph{Continuous sign corpora} contain parallel sequences of signs and spoken language.
 Available continuous sign corpora are extremely limited, containing 4-6 orders of magnitude fewer sentence pairs than similar corpora for spoken language machine translation \cite{arivazhagan2019massively}.
Moreover, while automatic speech recognition (ASR) datasets contain up to 50,000 hours of recordings \cite{pratap2020mls}, the largest continuous sign language corpus contain only 1,150 hours, and only 50 of them are publicly available \cite{dataset:hanke-etal-2020-extending}.
These datasets are usually synthesized \cite{dataset:databases2007volumes,dataset:Crasborn2008TheCN,dataset:ko2019neural,dataset:hanke-etal-2020-extending} or recorded in studio conditions \cite{dataset:forster2014extensions,cihan2018neural}, which does not account for noise in real-life conditions. Moreover, some contain signed interpretations of spoken language rather than naturally-produced signs, which may not accurately represent native signing since translation is now a part of the discourse event.

\paragraph{Availability}
Unlike the vast amount and diversity of available spoken language resources that allow various applications, 
signed language resources are scarce and currently only support translation and production.
Unfortunately, most of the signed language corpora discussed in the literature are either not available for use or available under heavy restrictions and licensing terms. Signed language data is especially challenging to anonymize due to the importance of facial and other physical features in signing videos, limiting its open distribution, and developing anonymization with minimal information loss, or accurate anonymous representations is a promising research problem.

\subsection{Sign Language Processing Tasks}\label{sec:modeling-apps}

The CV community has mainly led the research on SLP so far to focus on processing the visual features in signed language videos. As a result, current SLP methods do not fully address the linguistic complexity of signed languages.
We survey common SLP tasks and limitations of current methods by drawing on linguistic theories of signed languages.


\paragraph{Detection}
Sign language detection is the binary classification task to determine whether a signed language is being used or not in a given video frame. While recent detection models \cite{detection:borg2019sign,detection:moryossef2020real} achieve high performance, we lack well-annotated data that include interference and distractions with non-signing instances for proper evaluation.
A similar task in spoken languages is voice activity detection (VAD) \cite{sohn1999statistical,ramirez2004efficient}, the detection of when a human voice is used in an audio signal. However, as VAD methods often rely on speech-specific representations such as spectrograms, they are not always applicable to videos.

\paragraph{Identification}
Sign language identification classifies which signed language is being used in a given video automatically.
Existing works utilize the distribution of phonemes \cite{identification:gebre2013automatic} or activity maps in signing space \cite{identification:monteiro2016detecting} to identify the signed language in videos. However, these methods only rely on low-level visual features, while signed languages have several distinctive features on a linguistic level, such as lexical or structural differences \cite{mckee2000lexical, kimmelman2014information, ferreira1984similarities, shroyer1984signs} which have not been explored for this task.

\paragraph{Segmentation}
Segmentation consists of detecting the frame boundaries for signs or phrases in videos to divide them into meaningful units. Current methods resort to segmenting units loosely mapped to signed language units \cite{santemiz2009automatic, farag2019learning, bull2020automatic}, and does not leverage reliable linguistic predictors of sentence boundaries such as prosody in signed languages (i.e. pauses, sign duration, facial expressions, eye apertures) \cite{sandler2010prosody, ormel2012prosodic}.

\paragraph{Recognition}\label{sec:slr}
Sign language recognition (SLR) detects and label signs from a video, either on isolated  \cite{dataset:imashev2020dataset,dataset:sincan2020autsl} or continuous \cite{cui2017recurrent,camgoz2018neural,camgoz2020sign} signs. Though some previous works have referred to this as ``sign language translation'', recognition merely determines the associated label of each sign, without handling the syntax and morphology of the signed language \cite{padden1988interaction} to create a spoken language output.
Instead, SLR has often been used as an intermediate step during translation to produce glosses from signed language videos.

\paragraph{Translation}
Sign language translation (SLT) commonly refers to the translation of signed language to spoken language. 
Current methods either perform translation with glosses \cite{camgoz2018neural,camgoz2020sign,yin2020attention, yin-read-2020-better, moryossef2021data} or on pose estimations and sign articulators from videos \cite{dataset:ko2019neural, camgoz2020multi}, but do not, for instance, handle spatial relations and grounding in discourse to resolve ambiguous referents.

\paragraph{Production}\label{sec:models-slp}
Sign language production consists of producing signed language from spoken language and often use poses as an intermediate representation to overcome challenges in animation.
To overcome the challenges in generating videos directly, most efforts use poses as an intermediate representation, with the goal of either using computer animation or pose-to-video models to perform video production.
Earlier methods generate and concatenate isolated signs \cite{stoll2018sign,stoll2020text2sign}, 
while more recent methods \cite{saunders2020progressive,saunders2020everybody,pose:zelinka2020neural,xiao2020skeleton} autoregressively decode a sequence of poses from an input text.
Due to the lack of suitable automatic evaluation methods of generated signs, existing works resort to measuring back-translation quality, which cannot accurately capture the quality of the produced signs nor its usability in real-world settings. A better understanding of how distinctions in meaning are created in signed language may help develop a better evaluation method.

\section{Towards Including Signed Languages in Natural Language Processing}\label{sec:applications}

The limitations in the design of current SLP models often stem from the lack of exploring the linguistic possibilities of signed languages. We therefore invite the NLP community to collaborate with the CV community, for their expertise in visual processing, and signing communities and sign linguists, for their expertise in signed languages and the lived experiences of signers, in researching SLP.
We believe that first, the development of known tasks in the standard NLP pipeline to signed languages will help us better understand how to model them, as well as provide valuable tools for higher-level applications.
Although these tasks have been thoroughly researched for spoken languages, they pose interesting new challenges in a different modality. We also emphasize the need for real-world data to develop such methods, and a close collaboration with signing communities to have an accurate understanding of how signed language technologies can benefit signers, all the while respecting the Deaf community's ownership of signed languages.


\subsection{Building NLP Pipelines}\label{sec:applications-tools}
Although signed and spoken languages differ in modality, we argue that as both express the syntax, semantics, and pragmatics of natural languages, fundamental theories of NLP can and should be extended to signed languages. NLP applications often rely on low-level tools such as tokenizers and parsers, so we invite more research efforts on these core NLP tasks that often lay the foundation of other applications. We also discuss what considerations should be taken into account for their development to signed languages and raise open questions that should be addressed. 

\paragraph{Tokenization} 
The vast majority of NLP methods require a discrete input. To extend NLP technologies to signed languages, we must first and foremost be able to develop adequate tokenization tools that maps continuous signed language videos to a discrete, accurate representation with minimal information loss.
While existing SLP systems and datasets often use glosses 
as discrete lexical units of signed phrases, this poses three significant problems: (1) linear, single-dimensional glosses cannot fully capture the spatial constructions of signed languages, which downgrades downstream performance \cite{yin-read-2020-better}; (2) glosses are language-specific and requiring new glossing models for each language is impractical given the scarcity of resources; (3) glosses lack standard across corpora which limits data sharing and adds significant overhead in modeling. 

We thus urge the adoption of an \emph{efficient}, \emph{universal}, and \emph{standardized} method for tokenization of signed languages, all the while considering: how do we define lexical units in signed languages? \cite{johnston1999defining, johnston2010archive} To what degree can phonological units of signed languages be mapped to lexical units? Should we model the articulators of signs separately or together? What are the cross-linguistic phonological differences to consider? To what extent can ideas used in automatic speech recognition be applied to signed languages?



\paragraph{Syntactic Analysis}
Part-of-speech (POS) tagging and syntactic parsing are fundamental to understand the meaning of words in context. 
Yet, no such linguistic tools for automatic syntactic analyses exist. 
To develop such tools, we must first define to what extent POS tagging and syntactic parsing for spoken languages also generalize to signed languages - do we need a new set of POS and dependency tags for signed languages? How are morphological features expressed? What are the annotation guidelines to create datasets on syntax? Can we draw on linguistic theories to design features and rules that perform these tasks? Are there typologically similar spoken languages for some signed languages we can perform transfer learning with? 

\paragraph{Named Entity Recognition (NER)}
Recognizing named entities and finding relationships between them are highly important components in information retrieval and classification. 
Named entities in signed languages can be produced by a fingerspelled sequence, a sign, or even through mouthing of the name while the referent is introduced through pointing.
\citet{bleicken-etal-2016-using} attempt NER in German Sign Language (DGS) to perform anonymization, but only do so indirectly, by either performing NER on the gold DGS gloss annotations and German translations or manually on the videos. We instead propose NER in a fully automated fashion while considering, what are the visual markers of named entities? How are they introduced and referenced? How are relationships between them established?

\paragraph{Coreference Resolution}
Resolving coreference is crucial for language understanding. In signed languages, present referents, where the signer explicitly points to the entity in question, are relatively unambiguous. In contrast, non-present referents and classifiers are heavily grounded in the signing space, so good modeling of the spatial coherence in sign language is required. Evidence suggests that classic theoretical frameworks, such as discourse representation theory, may extend to signed languages \cite{steinbach2016drt}. We pose the following questions: to what extent can automatic coreference resolution of spoken languages be applied to signed languages? How do we keep track of referents in space? How can we leverage spatial relations to resolve ambiguity?

\paragraph{Towards Linguistically Informed and Multimodal SLP}

We highly encourage the collaboration of multimodal and SLP research communities to develop powerful SLP models informed by core NLP tools such as the ones discussed, all the while processing and relating information from both linguistic and visual modalities. On the one hand, theories and methods to reason multimodal messages can enhance the joint modeling of vision and language in signed languages. SLP is especially subject to three of the core technical challenges in multimodal machine learning \cite{baltruvsaitis2018multimodal}: \textbf{translation} - how do we map visual-gestural information to/from audio-oral and textual information?  \textbf{alignment} - how do we relate signed language units to spoken language units? \textbf{co-learning} - can we transfer high-resource spoken language knowledge to signed language? On the other hand, meaning in spoken languages is not only conveyed through speech or text but also through the visual modality. Studying signed languages can give a better understanding of how to model co-speech gestures, spatial discourse relations, and conceptual grounding of language through vision.

\subsection{Collect Real-World Data}\label{sec:applications-data}

Data is essential to develop any of the core NLP tools previously described, and current efforts in SLP are often limited by the lack of adequate data. We discuss the considerations to keep in mind when building datasets, challenges of collecting such data, and directions to facilitate data collection.

\paragraph{What is Good Signed Language Data?}
For SLP models to be deployable, they must be developed using data that represents the real world accurately. What constitutes an ideal signed language dataset is an open question,
we suggest including the following requirements: 
(1) a broad domain; (2) sufficient data and vocabulary size; (3) real-world conditions; (4) naturally produced signs; (5) a diverse signer demographic; (6) native signers; and when applicable, (7) dense annotations. 

To illustrate the importance of data quality during modeling, we first take as an example a current benchmark for SLP, the RWTH-PHOENIX-Weather 2014T dataset \cite{cihan2018neural} of German Sign Language, that does not meet most of the above criteria: it is restricted to the weather domain (1); contains only around 8K segments with 1K unique signs (2); filmed in studio conditions (3); interpreted from German utterances (4); and signed by nine Caucasian interpreters (5,6).
Although this dataset successfully addressed data scarcity issues at the time and successfully rendered results comparable and fueled competitive research, it does not accurately represent signed languages in the real world. On the other hand, the Public DGS Corpus \cite{dataset:hanke-etal-2020-extending} is an open-domain (1) dataset consisting of 50 hours of natural signing (4) by 330 native signers from various regions in Germany (5,6), annotated with glosses, HamNoSys and German translations (7), meeting all but two requirements we suggest. 

We train a gloss-to-text sign language translation transformer \cite{yin-read-2020-better} on both datasets. On RWTH-PHOENIX-Weather 2014T, we obtain \textbf{22.17} BLEU on testing; on Public DGS Corpus, we obtain a mere \textbf{3.2} BLEU. Although Transformers achieve encouraging results on RWTH-PHOENIX-Weather 2014T \cite{saunders2020progressive, camgoz2020multi}, they fail on more realistic, open-domain data. These results reveal that firstly, for real-world applications, we need more data to train such types of models, and secondly, while available data is severely limited in size, less data-hungry and more linguistically-informed approaches may be more suitable. This experiment reveals how it is crucial to use data that accurately represent the complexity and diversity of signed languages to precisely assess what types of methods are suitable, and how well our models would deploy to the real world.



\paragraph{Challenges of Data Collection}
Collecting and annotating signed data inline with the ideal requires more resources than speech or text data, taking up to 600 minutes per minute of an annotated signed language video \cite{dataset:hanke-etal-2020-extending}. Moreover, annotation usually requires a specific set of knowledge and skills, which makes recruiting or training qualified annotators challenging. Additionally, there is little existing signed language data in the wild that are open to use, especially from native signers that are not interpretations of speech. 
Therefore, data collection often requires significant efforts and costs of on-site recording as well. 

\paragraph{Automating Annotation}
To collect more data that enables the development of deployable SLP models, one useful research direction is creating tools that can simplify or automate parts of the collection and annotation process. One of the largest bottleneck in obtaining more adequate signed language data is the amount of time and scarcity of experts required to perform annotation. Therefore, tools that perform automatic parsing, detection of frame boundaries, extraction of articulatory features, suggestions for lexical annotations, and allow parts of the annotation process to be crowdsourced to non-experts, to name a few, have a high potential to facilitate and accelerate the availability of good data.

\subsection{Practice Deaf Collaboration}\label{sec:applications-deaf}

Finally, when working with signed languages, it is vital to keep in mind \emph{who} this technology should benefit, and \emph{what} they need.
Researchers in SLP must honor that signed languages belong to the Deaf community and avoid exploiting their language as a commodity \cite{bird-2020-decolonising}.

\paragraph{Solving Real Needs}

Many efforts in SLP have developed intrusive methods (e.g. requiring signers to wear special gloves), which are often rejected by signing communities and therefore have limited real-world value.
Such efforts are often marketed to perform ``sign language translation" when they, in fact, only identify fingerspelling or recognize a very limited set of isolated signs at best. These approaches oversimplify the rich grammar of signed languages, promote the misconception that signs are solely expressed through the hands, and are considered by the Deaf community as a manifestation of audism, where it is the signers who must make the extra effort to wear additional sensors to be understood by non-signers \cite{erard2017sign}. In order to avoid such mistakes, we encourage close Deaf involvement throughout the research process to ensure that we direct our efforts towards applications that will be adopted by signers, and do not make false assumptions about signed languages or the needs of signing communities. 

\paragraph{Building Collaboration}
Deaf collaborations and leadership are essential for developing signed language technologies to ensure they address the community's needs and will be adopted, and that they do not rely on misconceptions or inaccuracies about signed language \cite{harris2009research, kusters2017innovations}. 
Hearing researchers cannot relate to the deaf experience or fully understand the context in which the tools being developed would be used, nor can they speak for the deaf. Therefore, we encourage the creation of a long-term collaborative environment between signed language researchers and users, so that deaf users can identify meaningful challenges, and provide insights on the considerations to take, while researchers cater to the signers' needs as the field evolves. We also recommend reaching out to signing communities for reviewing papers on signed languages, to ensure an adequate evaluation of this type of research results published at ACL venues. There are several ways to connect with Deaf communities for collaboration: one can seek deaf students in their local community, reach out to schools for the deaf, contact deaf linguists, join a network of researchers of sign-related technologies\footnote{\url{https://www.crest-network.com/}}, and/or participate in deaf-led projects.

\section{Conclusions}

We urge the inclusion of signed languages in NLP. We believe that the NLP community is well-positioned, especially with the plethora of successful spoken language processing methods coupled with the recent advent of computer vision tools for videos, to bring the linguistic insight needed for better signed language models.
%
%
We hope to see an increase in both the interests and efforts in collecting signed language resources and developing signed language tools while building a strong collaboration with signing communities.


\section*{Acknowledgements}
We would like to thank Marc Schulder, Claude Mauk, David Mortensen, Chaitanya Ahuja, Siddharth Dalmia, Shruti Palaskar, Graham Neubig, and Becky Norton as well as the anonymous reviewers for their helpful feedback and insightful discussions.


\bibliographystyle{acl_natbib}
\bibliography{anthology,acl2021}


\end{document}